\documentclass[sn-mathphys-num,iicol]{sn-jnl}% Default with double column layout

\usepackage{graphicx}%
\usepackage{multirow}%
\usepackage{amsmath,amssymb,amsfonts}%
\usepackage{amsthm}%
\usepackage{mathrsfs}%
\usepackage[title]{appendix}%
\usepackage{xcolor}%
\usepackage{textcomp}%
\usepackage{manyfoot}%
\usepackage{booktabs}%
\usepackage{algorithm}%
\usepackage{algorithmicx}%
\usepackage{algpseudocode}%
\usepackage{listings}%
\usepackage{array}
\usepackage{siunitx}
\usepackage{stfloats}
\usepackage{url}
\usepackage{verbatim}
\usepackage{lettrine}
\usepackage{bm}
\usepackage{float}
\usepackage{natbib}
\usepackage[export]{adjustbox}
\DeclareMathOperator*{\argmax}{arg\,max}
\DeclareMathOperator*{\argmin}{arg\,min}

\begin{document}

\title[AFS-BM: Enhancing Model Performance through Adaptive Feature Selection with Binary Masking]{AFS-BM: Enhancing Model Performance through Adaptive Feature Selection with Binary Masking}
\linespread{0.95}

\author*[1]{\fnm{Mehmet Y.} \sur{Turali}}\email{yigit.turali@ug.bilkent.edu.tr}
\equalcont{These authors contributed equally to this work.}
\author[1]{\fnm{Mehmet E.} \sur{Lorasdagi}}\email{efe.lorasdagi@bilkent.edu.tr}
\equalcont{These authors contributed equally to this work.}


\author[1]{\fnm{Suleyman S.} \sur{Kozat}}\email{kozat@ee.bilkent.edu.tr}

\affil*[1]{\orgdiv{Department of Electrical and Electronics Engineering}, \orgname{Bilkent University}, \orgaddress{\city{Ankara}, \postcode{06800}, \country{Turkey}}}

\abstract{We study the problem of feature selection in general machine learning (ML) context, which is one of the most critical subjects in the field. Although, there exist many feature selection methods, however, these methods face challenges such as scalability, managing high-dimensional data, dealing with correlated features, adapting to variable feature importance, and integrating domain knowledge. To this end, we introduce the ``Adaptive Feature Selection with Binary Masking" (AFS-BM) which remedies these problems. AFS-BM achieves this by joint optimization for simultaneous feature selection and model training. In particular, we do the joint optimization and binary masking to continuously adapt the set of features and model parameters during the training process. This approach leads to significant improvements in model accuracy and a reduction in computational requirements. We provide an extensive set of experiments where we compare AFS-BM with the established feature selection methods using well-known datasets from real-life competitions. Our results show that AFS-BM makes significant improvement in terms of accuracy and requires significantly less computational complexity. This is due to AFS-BM's ability to dynamically adjust to the changing importance of features during the training process, which an important contribution to the field. We openly share our code for the replicability of our results and to facilitate further research.}

\keywords{Machine Learning, Feature Selection, Gradient Boosting Machines, Adaptive Optimization, Binary mask, High-Dimensional Datasets}



\maketitle

\section{Introduction}\label{Introduction}

The emergence of the digital era has enabled a significant increase in data across various domains, including genomics and finance. High-dimensional datasets are valuable in areas such as image processing, genomics, and finance due to the detailed information they provide \citep{capobianco2022highdimensional}. These datasets can contribute to the development of highly accurate models. This influx of data, particularly in the form of high-dimensional feature sets, brings both opportunities and challenges for the machine learning domain \citep{bellman1961adaptive}. High-dimensional data can reveal complex patterns, but it may also lead to the well-known “curse of dimensionality” \citep{bellman1961adaptive}. The large number of features can sometimes hide true patterns, leading to models that require significant computational resources prone to overfitting and are difficult to interpret \citep{bishop2006pattern}. 

While abundant features can provide rich information, not all features contribute equally to the final task. Some might be redundant, irrelevant, or even detrimental to the performance of the model if there is not enough data to learn from \citep{Farmanbar2016}. The challenge lies in identifying which features are essential and which can be discarded without compromising the accuracy of the model with limited data \citep{guyon2003introduction,Lajevardi2012}. This problem of feature selection is crucial for both regression and classification tasks; as an example, this happens when the feature dimensions are comparable to the training dataset size or for non-stationary environments, i.e., when we do not have enough data to learn which features are relevant due to changing statistics.

Here, we introduce ``Adaptive Feature Selection with Binary Masking" (AFS-BM), a novel method designed for feature selection in high-dimensional datasets and scenarios characterized by non-stationarity. This method is particularly adept at handling cases with insufficient data for reliably identifying relevant features, a challenge it addresses through the  integration of a binary mask. This binary mask, represented as a column vector with ``1"s for active features and ``0"s for inactive ones, plays a pivotal role in our joint optimization framework, seamlessly combining feature selection and model training, which is normally an NP hard optimization problem. AFS-BM performs this by joint adaptation of the binary mask concurrently with the model parameter optimization. This joint and dynamic optimization enables us to fine-tune the feature set concurrently and enhance model training, resulting in a more agile and efficient identification of relevant features. Consequently, it significantly elevates the predictive accuracy of the underlying model. Our comprehensive experimentation across renowned competition datasets demonstrates the robustness and effectiveness of our method. Compared to widely used feature selection techniques, AFS-BM consistently yields substantial performance improvements. Furthermore, we also openly share our code\footnote{\url{https://github.com/YigitTurali/AFS_BM-Algorithm}} to facilitate further research and replicability of our results.

Our main contributions to the literature can be summarized as follows:

\begin{itemize}
    \item We introduce AFS-BM that uniquely combines binary mask feature selection with joint optimization. This integration allows for dynamic feature selection, actively improving model accuracy by focusing on the most relevant features and simultaneously reducing noise by eliminating less significant ones. 
    \item This binary mask dynamically refine the feature set, playing a crucial role in improving the computational efficiency of the approach.
    \item  Our algorithm ensures a critical balance between feature selection and maintaining model accuracy by using a well-tuned and iterative approach that evaluates and prunes features based on their final impact on performance, thereby ensuring efficiency without compromising predictive precision.
    \item We demonstrate the significant performance improvements over traditional feature selection techniques, particularly in its application to GBMs and NNs.\footnote{The rest of the paper is organized as follows: Section \ref{related_work} outlines the current literature of the filter, wrapper, embedded and adaptive feature selection methods. Section \ref{sec:prob_desc} presents the mathematical background and problem description with current feature selection methods. Section \ref{sec:our_algs} introduces our novel feature selection structure, detailing the algorithms and their underlying principles. Section \ref{sec:experiments} showcases our experimental results, comparing our approach with established feature selection techniques. Finally, Section \ref{sec:conclusion} offers a summary of our findings and conclusions.}
\end{itemize}

\section{Related Work}\label{related_work}

Several established methods for feature selection exist. Filter methods such as the Chi-squared test, mutual information \citep{peng2005feature}, and correlation coefficients assess features based on their inherent statistical characteristics \citep{kira1992practical}. Wrapper methods, including techniques such as sequential forward selection and sequential backward elimination, employ specific machine learning algorithms to evaluate varying feature subsets \citep{Rida2016}. Embedded methods, represented by techniques such as LASSO and decision tree-based approaches, integrate feature selection directly into the model training process \citep{tibshirani1996regression, breiman2001random}. However, these methods often depend on static analysis and predetermined algorithms, which can be a limitation. Specifically, methods like wrappers and embedded approaches may require significant computational resources, making them challenging to apply to large datasets \citep{saeys2007review}. While widely used, these methods often struggle with high-dimensional datasets and lack the flexibility to adapt to the evolving requirements of complex models \citep{Atan2019}. In contrast, our method employs a binary mask coupled with iterative optimization, enabling dynamic adaptation to the learning patterns of a model. Since this approach jointly and dynamically adjust to the changing importance of features during the training process, it enhances the precision of feature selection and ensures better performance in challenging scenarios involving high-dimensional or evolving datasets. Additionally, in the context of evolving datasets, where data patterns and relationships may shift over time, the flexibility of the binary mask to adapt to these changes ensures that the model remains robust and relevant. It can adjust to new patterns or discard previously relevant features that become obsolete, maintaining its efficacy in dynamically changing environments. This adaptability is crucial for maintaining high performance over time, especially in real-world applications where data is not static and can evolve. 

In recent years, there has been a shift towards adaptive feature selection methods. These methods focus on adapting to the learning behavior of the model. An example of this is the Recursive Feature Elimination (RFE) for Support Vector Machines (SVMs), where features are prioritized based on their influence on the SVM’s decision-making \citep{Subasi2015, Ruszczak2024}. Tree-based models, such as Random Forests and GBMs, also provide feature importance metrics derived from their tree structures \citep{chen2016xgboost}. These existing methods, while adaptive, are not always optimal for high-dimensional and non-stationary data, and they are designed mainly for traditional machine learning models and might not be optimal for different architectures such as for NNs \citep{goodfellow2016deep}. In contrast, AFS-BM demonstrates superior adaptability and efficiency in handling such challenges. Its iterative refinement of feature selection through a binary mask is effective, offers more precise feature selection and making it a more versatile and robust solution because it continuously evaluates and adjusts which features are most predictive, thus improving focus of the model on truly relevant data for NNs and other advanced machine learning algorithms. 

\section{Problem Statement}\label{sec:prob_desc}
Vectors in this manuscript are represented as column vectors using bold lowercase notation, while matrices are denoted using bold uppercase letters.\footnote{For a given vector $ \bm{x} $ and a matrix $\bm{X}$, their respective transposes are represented as $ \bm{x}^T $ and $ \bm{X}^T $. The symbol $ \odot $ denotes the Hadamard product, which signifies an element-wise multiplication operation between matrices. For any vector $ \bm{x} $, $ x_i $ represents the $ i^{th} $ element. For a matrix $ \bm{X} $, $ X_{ij} $ indicates the element in the $ i^{th} $ row and $ j^{th} $ column. The operation $ \sum(\cdot) $ calculates the sum of the elements of a vector or matrix. The L1 norm of a vector \( \bm{x} \) is defined as  $||\bm{x}||_1 = \sum_{i} |x_i|,$ where the summation runs over all elements of the vector. The number of elements in set \( S \) is given by \( |S| \).}

We address the feature selection problem in large datasets and explore the challenges associated with time series forecasting and classification tasks. We then examine GBMs and RFEs, concluding with a discussion of widely used feature selection methods in the machine learning literature, such as cross-correlation and mutual information.

We study adaptive feature selection in the context of online learning for prediction/forecasting/classification of non-stationary data, where we adaptively learn the most relevant features. Given a vector sequence \( \bm{x}_{1:T} = \{\bm{x}_t\}_{t=1}^{T} \), where \( T \) denotes the sequence length and \( \bm{x}_t \in \mathbb{R}^{M} \) is the feature vector at time \( t \) where this input can include features from the target sequence (endogenous) and auxiliary (exogenous) features such as weather or time of the day. The corresponding target output for \( \bm{x}_{1:T} \) is \( \bm{y}_{1:T} = \{\bm{y}_t\}_{t=1}^{T} \), with \( \bm{y}_t \in \mathbb{R}^{C} \) being the desired output vector at time \( t \), $C$ where represents the number of components or dimensions in the desired output vector.

In the online learning setting, the goal is to estimate \( \bm{y}_t \) using only the inputs observed up to time \( t \) as \( \bm{\hat{y}}_t = f_t(\bm{y}_{1},\ldots,\bm{y}_{t-1},\bm{x}_1,\ldots,{\bm{x}_t}; \bm{\theta}_t) \), where \( f_t \) is a dynamic nonlinear function parameterized by \( \bm{\theta}_t \). After each observation of \( \bm{y}_t \), we compute a loss \( \mathcal{L}(\bm{y}_t, \bm{\hat{y}}_t) \) and update the algorithm parameters in real-time. As an example, for this paper, such as the mean squared error (MSE), is computed over the sequence:

\begin{equation}
\mathcal{L}_{\text{MSE}} = \frac{1}{T} \sum_{t=1}^{T} \bm{e}_t^T \bm{e}_t,
\end{equation}

\noindent where $ \bm{e}_t = \bm{y}_t - \bm{\hat{y}}_t $ denotes the error vector at time $ t $. Other metrics, such as the mean absolute error (MAE), can also be considered since our method is generic and extends to other loss functions.

Additionally, we address adaptive feature selection for classification tasks. Given a set of feature vectors $ \bm{X} = [\bm{x}_1,\ldots,\bm{x}_N]^T $, where $ N $ is the total number of samples and $ \bm{x}_i \in \mathbb{R}^{M} $ is the feature vector for the $ i^{th} $ sample, this input can encompass both primary features and auxiliary information. For the dataset \( \bm{X} \), the corresponding target labels are given by \( \bm{Y} = [\bm{y}_1,\ldots,\bm{y}_N]^T \). Here, $\bm{y}_i \in \mathbb{R}^{|C|} $, where each class is denoted by \( c \) which belongs to the set \( C \) and $ y_{i,c}  = p$ represents the true probability of class  \( c \) for the \( i^{th} \) sample. The goal is to predict the class label $ \hat{y}_i =  \argmax_{c} \hat{y}_{i,c} $ using the feature vector $ \bm{x}_i $ as $ \bm{\hat{y}}_i = f(\bm{x}_i; \bm{\theta}) $, where $ f $ is a nonlinear classifier function parameterized by $  \bm{\theta} $ and the corresponding target label probabilities are given by \( \bm{\hat{Y}} =[\bm{\hat{y}}_1,\ldots,\bm{\hat{y}}_N]^T \) where \( \bm{\hat{y}}_i\in [0,1]^{|C|} \). Here, $\bm{\hat{y}}_i$  represents the probability vector of all class labels for the \( i^{th} \) sample, where each class is denoted by \( c \) and belongs to the set \( C \). Upon observing the true probability vector of labels $ \bm{y}_i$, we incur a loss $ \mathcal{L}(\bm{y}_i, \bm{\hat{y}}_i) $ and adjust the model based on this loss. The performance of the model is assessed by classification accuracy or other relevant metrics. A commonly used metric is the cross-entropy loss $\mathcal{L}_{\text{CE}} = -\frac{1}{N} \sum_{i=1}^{N} \sum_{c=1}^{C} y_{i,c} \log(\hat{y}_{i,c})$ where ${y}_{i,c} $ is the true probability of class $ c $ for the $ i^{th} $ sample, and $ \hat{{y}}_{i,c} $ is the predicted probability. Other metrics, such as the F1-score or Area Under the Curve (AUC), can also be used.
 
As the feature space expands, the risk of overfitting increases, especially when the number of features $M$ approaches the number of samples $N$ \citep{bellman1961adaptive}, often referred to as the “Curse of Dimensionality”. For example, non-stationarity can arise in regression from evolving relationships between variables \citep{Junejo2013}. In classification, non-stationarity can manifest as shifting class boundaries due to changing class distributions \citep{Aguiar2023}.

Among the various feature selection techniques developed, some of the most effective and widely used methods include GBMs for feature importance \citep{zhang2022forest}, RFE, and Greedy Methods. These methods provide robust frameworks for identifying significant features within large datasets. Building upon these foundational techniques, we propose a novel method described in the following sections.

\section{Adaptive Feature Selection with Binary Masking}\label{sec:our_algs}
Our method, AFS-BM, continually refines its choice of features by utilizing a binary mask. We first introduce the “Model Optimization Phase” (\ref{sec:model_opt}), detailing how features are initially selected and utilized for model training. We then continue to the “Masked Optimization \& Feature Selection Phase” (\ref{sec:mask_opt}), where we describe the process of refining the feature set and optimizing the binary mask. The combined approach ensures a systematic and adaptive feature selection for improved model performance.

Consider a dataset at the start of the algorithm, which is denoted as $\mathcal{D} = \{(\bm{x}_i, y_i)\}_{i=1}^{N+P}$ where $N+P$ is the number of samples in the dataset since we will divide the dataset in further. Here, $\bm{x}_i \in \mathbb{R}^M$ represent the feature vector for sample $i$, and $y_i$ are the corresponding target value in the dataset $\mathcal{D}$, respectively. The set of feature vectors is defined as $ \bm{X} = [\bm{x}_1, \ldots,\bm{x}_N]^T $ and the target vector defined as $\bm{y} = [{y}_1, \ldots,{y}_N]^T $. To extract important features and eliminate redundant features, here, we define a binary mask, \( \bm{z} \) as $\bm{z} \in \{0, 1\}^M$ where \( M \) is the total number of features. The binary nature ensures that a feature is either selected (1) or not (0) where the set of feature vectors $\bm{X}$ are modified by the mask with the Hadamard product, such as:

\begin{equation}
    \bm{X}_{modified} = \bm{X} \odot \bm{z}.
\end{equation}

\noindent Moreover, the binary mask \( \bm{z} \) can be seen as a constraint on the feature space. Therefore, the primary objective is to minimize the loss function while optimizing this mask, which closely approximates the true target values and selects the best features by using the minimum number of features as possible. This problem can be represented as:
\begin{equation}
    \min_{\bm{z}, \bm{\theta}} \mathcal{L}(\bm{y}, F(\bm{X} \odot \bm{z},\bm{\theta})) + \frac{||\bm{z}||_1}{M} ,
\end{equation}\label{alg:opt_problem}

\noindent as subject to \( \bm{z} \in \{0, 1\}^M \). Note that (\ref{alg:opt_problem}) does not have a closed-form solution since it includes integer optimization \citep{sun2018distributed}. Hence, we propose an iterative algorithm that overcomes the challenges posed by the integer optimization in (\ref{alg:opt_problem}). Our algorithm leverages the subgradient methods \citep{sun2018distributed}, and introduces a relaxation technique for the binary constraints on \( \bm{z} \). In each iteration, the algorithm refines the approximation of \( \bm{z} \) and updates \( \bm{\theta} \) based on the current estimate. The iterative process continues until convergence, i.e., when the change in the objective function value between consecutive iterations falls below a predefined threshold.

As a solution to the optimization problem presented in (\ref{alg:opt_problem}), our algorithm AF-BM detailed in Algorithm \ref{alg:AFSBM} iteratively performs both binary mask and loss optimization. The binary mask acts as a dynamic filter, allowing the algorithm to select or discard features during the learning process actively. This ensures that only the most relevant features are used, thereby improving the accuracy of the model. The iterative optimization refines both the parameters of the model and the binary mask simultaneously, ensuring that the feature selection process remains aligned with the learning objectives of the model. The algorithm continues its iterative process until the binary mask remains unchanged between iteration cycles, providing a robust and efficient solution to the problem.

The AFS-BM algorithm begins by initializing slack thresholds based on the cross-validation results, which are represented by the variables $\mu$ and $\beta$, as well as a positive real number $\Delta \mathcal{L}$ in line \ref{lst:line:ln1} of Algorithm \ref{alg:AFSBM}. At the start of the algorithm, specifically at iteration \( k = 0 \), two distinct datasets are used. The first dataset, denoted by \( \mathcal{D}^{(0)} \), is used for model optimization and is defined as  $\mathcal{D}^{(0)} = \{(\bm{x}_i, y_i)\}_{i=1}^N$. The second dataset, represented by \( \mathcal{D}_m^{(0)} \), is employed for the masked feature selection process and is given by 
$\mathcal{D}_m^{(0)} = \{(\bm{x}_i^{(m)}, y_i^{(m)})\}_{i=1}^P$. Both the definitions of  \( \mathcal{D}_m^{(0)} \) and \( \mathcal{D}_m^{(0)} \) are given in line \ref{lst:line:ln2}. In the notation, the superscript in \( \mathcal{D}^{(0)} \) and \( \mathcal{D}_m^{(0)} \) indicates the iteration number, which in this case is the initial iteration, i.e., 0 and the subscript ``$m$" in \( \mathcal{D}_m \) signifies that this dataset is specifically used for ``masked" feature selection. Here, $N$ and $P$ are the numbers of samples in the datasets. $\bm{x}_i,\bm{x}_i^{(m)} \in \mathbb{R}^M$ represent the feature vectors for sample $i$, and $y_i, y_i^{(m)}$ are the corresponding target values in model optimization and masked feature selection datasets, respectively. The set of feature vectors is defined as $ \bm{X}^{(0)} = [\bm{x}_1,\ldots,\bm{x}_N]^T $ and $ \bm{X}^{(0)}_m = [\bm{x}_1^{(m)},\ldots,\bm{x}_P^{(m)}]^T$ in line \ref{lst:line:ln3}. The process begins with the initialization of the binary mask vector in line \ref{lst:line:ln4}, $\bm{z}^{(0)} \in \{0, 1\}^M$, with all entries set to one, indicating the inclusion of all features. Our goal is to train a model and optimize a binary mask accordingly, $F_k(\bm{x}_i,\bm{\theta})$ and $\bm{z}$, that selects best features at the $k$-th iteration. 

\subsection{Model Optimization Phase of AFS-BM} \label{sec:model_opt}
In this phase, the focus is on using a binary mask from a prior iteration to identify the feature subset, which subsequently results in the formation of a masked dataset. The main objective is to fine-tune the model by reducing a given loss function while keeping the mask unchanged. After this optimization, a test loss is calculated and compared with a benchmark loss calculated on a separate masked dataset. This threshold enables the upcoming feature selection phase, ensuring the adaptive feature extraction.

The main loop of the algorithm continues until a stopping criterion, defined by $\beta \neq 0$, is met at line \ref{lst:line:ln5}. Within this loop, before the \(k\)-th feature selection iteration, the feature subset is determined by the binary mask \(\bm{z}^{(k-1)}\), which is optimal at iteration \(k-1\) concerning the minimization of the loss function from the previous iteration. This means that the mask \(\bm{z}^{(k-1)}\) was found to provide the best feature subset that resulted in the lowest loss or error for the model during the \(k-1\) iteration. First, at line \ref{lst:line:ln6}, the algorithm initializes a model $F_{k}(\bm{X}^{(k)}, \bm{\theta})$, and the $\bm{z}^{(k-1)}$ masks the feature vector and generate $\bm{x}_i^{masked}$ for the model training in line \ref{lst:line:ln7} consecutively $\bm{x}_i^{masked} = \bm{x}_i \odot \bm{z}^{(k-1)}$ which resulted in the following dataset in line \ref{lst:line:ln8}, $\mathcal{D}^{(k)} = \{(\bm{x}_i^{masked}, y_i)\}_{i=1}^N$. Hence, $ \bm{X}^{(k)} = [\bm{x}^{masked}_1,\ldots,\bm{x}^{masked}_N]^T$,where $ \bm{x}^{masked}_i \in \mathbb{R}^{M} $ is the feature vector for the $ i^{th} $ sample. Training involves minimizing the loss function $\mathcal{L}(\bm{y}, F_k(\bm{X}^{(k)},\bm{\theta}))$ with respect to the parameters of the model, $\bm{\theta}$ since the mask $\bm{z}^{(k-1)}$ is constant at this step which can be seen on line \ref{lst:line:ln10}:
\begin{equation}
    \bm{\hat {\theta}}_k = \argmin_{\bm{\theta}}\mathcal{L}(\bm{y}, F_k(\bm{X}^{(k)},\bm{\theta})).
\end{equation}
After optimizing the model, we determine the test loss value using the current best parameters, denoted as $\bm{\hat {\theta}}_k$. This loss value is then defined as $\mathcal{L}_{th}$ and will serve as a reference point or threshold for the subsequent phase of the algorithm. To calculate $\mathcal{L}_{th}$, we use the masked feature selection dataset $\mathcal{D}_m$ as a test set for the model optimization phase of the AFS-BM algorithm. First, we mask the feature vector with the most recent optimal mask $\bm{z}^{(k-1)}$ from the previous iteration and generate ${\bm{x}_i^{(m)}}^{masked}$. Then, we update $\mathcal{D}_m^{(k-1)}$ to calculate the $\mathcal{L}_{th}$ from line \ref{lst:line:ln11} to \ref{lst:line:ln14}:
\begin{align} \label{eq:after_train_calc} 
    &{\bm{x}_i^{(m)}}^{masked} = \bm{x}_i^{(m)} \odot \bm{z}^{(k-1)},\\
    &\mathcal{D}_m^{(k-1)} = \{({\bm{x}_i^{(m)}}^{masked}, y_i^{(m)})\}_{i=1}^P,\\
    &\bm{X}_{m}^{(k-1)} = [{\bm{x}_1^{(m)}}^{masked},\ldots,{\bm{x}_P^{(m)}}^{masked}]^T  ,\\
    &\mathcal{L}_{th} = \mathcal{L}(\bm{y}^{(m)}, F_k(\bm{X}_{m}^{(k-1)},\bm{\hat{\theta}}_k)),
\end{align}
\noindent where we define a test set of feature vectors at step $(k-1)$ as $ \bm{X}_{m}^{(k-1)} = [{\bm{x}_1^{(m)}}^{masked},\ldots,{\bm{x}_P^{(m)}}^{masked}]^T$, where $ {\bm{x}_i^{(m)}}^{masked} \in \mathbb{R}^{M} $ is the masked feature vector for the $ i^{th} $ sample of masked feature selection dataset. After calculating the threshold loss $\mathcal{L}_{th}$, we proceed with the masked feature selection phase.

\subsection{Masked Optimization \& Feature Selection Phase of AFS-BM} \label{sec:mask_opt}
In this phase, the AFS-BM algorithm undergoes a thorough feature extraction via mask optimization after training. It sets a slack variable to guide the optimization duration and uses a temporal mask to evaluate feature relevance. By iteratively masking out features and observing the impact on the loss of the model, the algorithm differentiates between essential and redundant features. This iterative process continues until the feature set stabilizes, ensuring the model is equipped with the most significant features for the minimum loss on the given dataset.

After training, the mask optimization and feature selection process starts. In this process we use the slack variable $\mu \in \mathbb{Z}^+$ to stop the mask optimization process, which is initialized in line \ref{lst:line:ln1} and adjusted based on cross-validation and only affects the algorithm’s computation time of the algorithm. The mask optimization phase starts with initializing a temporal mask $\bm{\hat{z}}^{(k-1)}$ that copies the binary mask from the previous iteration. We define the temporal mask $\bm{\hat{z}}^{(k-1)}$ at line \ref{lst:line:ln15} as:
\begin{equation}
\bm{\hat{z}}^{(k-1)} \gets \bm{z}^{(k-1)}. 
\end{equation}

\noindent As long as $\mu \neq 0$, which is met at line \ref{lst:line:ln17}, we continue with the index selection procedure. During this phase, the algorithm selects a unique index $i$ from a set of available indices $S$ based on a Gaussian distribution from lines \ref{lst:line:ln19} to \ref{lst:line:ln20}. This index selection procedure is described as follows:

Let $\mathcal{S} ={\{1, \ldots, M\}} $ be a set of indices with dimension $\dim{\mathcal{S}} = M$. An index $i$ is selected from $\mathcal{S}$ such that $i \in \mathcal{S}$. The selection of index \( i \) from \( \mathcal{S} \) is based on a uniform distribution. This probabilistic mechanism enhances the adaptiveness and diversity of the feature selection process. Once chosen, $i$ cannot be reselected, preserving the uniqueness of the selection. This is expressed as $i \in \mathcal{S}, \quad i \notin \{j \mid j \text{ has been previously selected}$.

\noindent Upon selecting index $i$, we set corresponding element of the temporal mask ${\hat{z}_i}^{(k-1)}$ to 0 as at line \ref{lst:line:ln22}, $\hat{z}_i^{(k-1)} \gets 0$.

\noindent With the temporal mask, $\bm{\hat{z}}^{(k-1)}$, we generate temporal feature vectors ${\bm{\hat{x}}_i^{(m)^{masked}}}$, a feature vector set $\bm{\hat{X}}_{m}^{(k)}$ and a dataset $\mathcal{\hat{D}}_m^{(k)}$ from lines \ref{lst:line:ln23} to \ref{lst:line:ln26}:

\begin{align}
    &{\bm{\hat{x}}_i^{(m)^{masked}}} = \bm{x}_i^{(m)} \odot \bm{\hat{z}}^{(k-1)},\\
    &\mathcal{\hat{D}}_m^{(k)} = \{({\bm{\hat{x}}_i^{(m)^{masked}}}, y_i^{(m)})\}_{i=1}^P,\\
    &\bm{\hat{X}}_{m}^{(k)} = [\bm{\hat{x}}_1^{(m)^{masked}},\ldots,\bm{\hat{x}}_N^{(m)^{masked}}]^T,  \\
    &\mathcal{L}_{mask} = \mathcal{L}(\bm{y}^{(m)},F_k(\bm{\hat{X}}_{m}^{(k)},\bm{\hat{\theta}}_k)),
\end{align}

\noindent and $\bm{\hat{X}}_{m}^{(k)} = \bm{X}_{m}^{(k-1)} \setminus \{i\}$ where \( \bm{X}_{m}^{(k-1)} \setminus \{i\} \) denotes the feature set without feature \( i \). We define the $\mathcal{L}_{mask}$ as $\mathcal{L}_{mask} = \mathcal{L}(\bm{y}^{(m)}, F_k(\bm{X}_{m}^{(t-1)}\setminus \{i\},\bm{\hat{\theta}}_t))$.

\noindent With the generated temporal feature vectors using the updated temporal mask and the algorithm calculates a new loss $\mathcal{L}_{mask}$ in line \ref{lst:line:ln27}. Moreover, we define that a feature \( i \) is said to be relevant if and only if $\frac{\mathcal{L}(\bm{y}^{(m)}, F_k(\bm{X}_{m}^{(k-1)}\setminus \{i\},\bm{\hat{\theta}}_k)) - \mathcal{L}(\bm{y}^{(m)}, F_k(\bm{X}_{m}^{(k-1)},\bm{\hat{\theta}}_k)))}{\mathcal{L}(\bm{y}^{(m)}, F_k(\bm{X}_{m}^{(k-1)},\bm{\hat{\theta}}_k)))} \leq \Delta \mathcal{L}$ or $\frac{\mathcal{L}_{mask}- \mathcal{L}_{th}}{\mathcal{L}_{th}} \leq \Delta \mathcal{L}$, where \( \bm{X}_{m}^{(k-1)} \setminus \{i\} \) denotes the feature set without feature \( i \) and $\Delta \mathcal{L}$ denotes a predefined threshold which is also initialized at line \ref{lst:line:ln1} and adjusted by cross-validation and only affects the sensitivity of determining feature’s relevance. The mask’s state is retained if the feature is relevant, i.e., the loss remains unchanged or changes by less than a predefined threshold, $\Delta \mathcal{L}$ as ${\hat{z}}^{(k-1)}_i =
\begin{cases}
0, & \text{if } \frac{\mathcal{L}_{mask}- \mathcal{L}_{th}}{\mathcal{L}_{th}} \leq \Delta \mathcal{L} \\
1, & \text{otherwise} .
\end{cases}$

\noindent If the loss does not exceed the threshold, we update $\mathcal{L}_{\text{th}}$ to $\mathcal{L}_{\text{mask}}$. Otherwise, we decrement $\mu$ and move to the next randomly chosen index. Let $\delta \mathcal{L}$ denote the relative change in the loss function when feature $i$ is removed, i.e., when ${\hat{z}}^{(k-1)}_i$ is set to 0. Formally, $\delta \mathcal{L} = \frac{\mathcal{L}_{mask}- \mathcal{L}_{th}}{\mathcal{L}_{th}}$. Given the threshold $\Delta \mathcal{L}$, a feature $i$ is deemed irrelevant if $\delta \mathcal{L} \leq \Delta \mathcal{L}$. This implies that the performance of the model does not degrade by more than $\Delta \mathcal{L}$ when feature $i$ is removed. Therefore, for any feature $i$ for which $\delta \mathcal{L} \leq \Delta \mathcal{L}$, the binary mask ensures that ${\hat{z}}^{(k-1)}_i$, ensuring that only features that significantly contribute to the performance of the model are retained. This process is shown in from lines \ref{lst:line:ln28} to \ref{lst:line:ln35}. The mask optimization concludes when $\mu = 0$. Once the mask optimization phase concludes, at line \ref{lst:line:ln36}, we then employ the optimized mask to eliminate redundant features:

\begin{align}
    \textbf{DeleteColumns}(\bm{X}^{(k)},\bm{X}_m^{(k-1)},\bm{\hat{z}}^{(k-1)}) = \\
    \bm{X}^{(k+1)},\bm{X}_m^{(k)}, \bm{z}^{(k)}.
    \label{eq:del_cols}
\end{align}

\noindent The function, $\textbf{DeleteColumns}(\cdot,\cdot,\cdot)$, removes feature columns from the datasets $\bm{X}^{(k)}$, $\bm{X}_m^{(k-1)}$ and $\bm{z}^{(k-1)}$ based on the temporal binary mask $\bm{\hat{z}}^{(k-1)}$. The mask \( \bm{z} \) is stable if:
\[ \bm{z}^{(k)} = \bm{z}^{(k-1)} ,\]
for a predefined number of consecutive iterations $\beta$, which is again adjusted by cross-validation and only affects the computation time of the algorithm. If the mask remains unchanged for a predefined number of consecutive iterations, represented by $\beta$, the algorithm determines that it reached an optimal set of features and either proceeds to the next $(k+1)$-th model optimization accordingly or terminates in between lines \ref{lst:line:ln37} and \ref{lst:line:ln40}.

The complete description of the algorithm can be found in Algorithm \ref{alg:AFSBM}.
\begin{algorithm}[!]
\small
\caption{Adaptive Feature Selection with Binary Masking (AFS-BM)}\label{alg:AFSBM}
\begin{algorithmic} [1]
\fontsize{9}{11}
\State Initialize $\mu,\beta \in \mathbb{Z}^{+}; \Delta \mathcal{L} \in \mathbb{R}^{+} $ \label{lst:line:ln1}  %\Comment{Initialize Slack Thresholds}
\State $\mathcal{D}^{(0)} = \{(\bm{x}_i, y_i)\}_{i=1}^N, \mathcal{D}_m^{(0)} = \{(\bm{x}_i^{(m)}, y_i^{(m)})\}_{i=1}^P$ \label{lst:line:ln2}
\vspace{0.5mm}%\Comment{Initialize Datasets}
\State $ \bm{X}^{(0)} = [\bm{x}_1,\ldots,\bm{x}_N]^T,\bm{X}_m^{(0)} = [\bm{x}_1,\ldots,\bm{x}_P]^T $ \label{lst:line:ln3}
\vspace{0.5mm}
\State $\bm{z}^{(0)} \in \{0, 1\}^M$ \label{lst:line:ln4}
\While{stop criterion $\beta \neq 0$} \label{lst:line:ln5}
    \State Initialize $F_k(\bm{X}^{(k)},\bm{\theta})$ \label{lst:line:ln6}
    \State $\bm{x}_i^{masked} = \bm{x}_i \odot \bm{z}^{(k-1)}$ \label{lst:line:ln7}
    \vspace{0.5mm}
    \State $\mathcal{D}^{(k)} = \{(\bm{x}_i^{masked}, y_i)\}_{i=1}^N$ \label{lst:line:ln8}
    \vspace{0.5mm}
    \State $ \bm{X}^{(k)} = [\bm{x}^{masked}_1,\ldots,\bm{x}^{masked}_N]^T $ \label{lst:line:ln9}
    \vspace{0.5mm}
    \State $\bm{\hat {\theta}}_k = \argmin_{\bm{\theta}}\mathcal{L}(\bm{y}, F_k(\bm{X}^{(k)},\bm{\theta}))$  \label{lst:line:ln10}
    \vspace{1mm}
    \State ${\bm{x}_i^{(m)}}^{masked} = \bm{x}_i^{(m)} \odot \bm{z}^{(k-1)}$ \label{lst:line:ln11} 
    \vspace{0.5mm}
    \State $\mathcal{D}_m^{(k-1)} = \{({\bm{x}_i^{(m)}}^{masked}, y_i^{(m)})\}_{i=1}^P$ \label{lst:line:ln12}
    \vspace{0.5mm}
    \State $ \bm{X}_m^{(k-1)} = [\bm{x}_1^{(m)},\ldots,\bm{x}_P^{(m)}]^T $ \label{lst:line:ln13}
    \vspace{0.5mm}
    \State $\mathcal{L}_{th} = \mathcal{L}(\bm{y}^{(m)}, F_k(\bm{X}_{m}^{(k-1)},\bm{\hat{\theta}}_k))$ \label{lst:line:ln14}
    \vspace{0.5mm}
    \State Define $\bm{\hat{z}}^{(k-1)} \gets \bm{z}^{(k-1)}$ \label{lst:line:ln15}
    \State Initialize set of available indices: $\mathcal{S}$ \label{lst:line:ln16}
    \While{stop criterion $\mu \neq 0$} \label{lst:line:ln17}
        \State Randomly select $i$ from $\mathcal{S}$  \label{lst:line:ln19}
        \vspace{0.5mm}
        \State Remove $i$ from $\mathcal{S}$ \label{lst:line:ln20}
        \State $\hat{z}_i^{(k-1)} = 0$ \label{lst:line:ln22}
        \vspace{0.5mm}
        \State ${\bm{\hat{x}}_i^{(m)^{masked}}} = \bm{x}_i^{(m)} \odot \bm{\hat{z}}^{(k-1)}$ \label{lst:line:ln23}
        \vspace{0.5mm}
        \State $\mathcal{\hat{D}}_m^{(k)} = \{({\bm{\hat{x}}_i^{(m)^{masked}}}, y_i^{(m)})\}_{i=1}^P$ \label{lst:line:ln24}
        \vspace{0.5mm}
        \State $ \bm{\hat{X}}_m^{(t)} = [\bm{\hat{x}}_1^{(m)^{masked}},\ldots,\bm{\hat{x}}_P^{(m)^{masked}}]^T$ \label{lst:line:ln25}
        \vspace{0.5mm}
        \State $ \bm{X}_{m}^{(k-1)} \setminus \{i\} = \bm{\hat{X}}_{m}^{(k)} $ \label{lst:line:ln26}
        \vspace{0.5mm}
        \State $\mathcal{L}_{mask} = \mathcal{L}(\bm{y}^{(m)}, F_k(\bm{X}_{m}^{(t-1)}\setminus \{i\},\bm{\hat{\theta}}_t))$ \label{lst:line:ln27}
        \vspace{0.5mm}
        \If{$ \frac{\mathcal{L}_{mask}- \mathcal{L}_{best}}{\mathcal{L}_{best}} \leq \Delta \mathcal{L}$} \label{lst:line:ln28}
        \vspace{0.5mm}
            \State $\hat{z}_i^{(k-1)} \gets 0$ \label{lst:line:ln29} 
            \vspace{0.5mm}
            \State $\mathcal{L}_{th} \gets \mathcal{L}_{mask}$ \label{lst:line:ln30}
        \Else
        \vspace{0.5mm}
            \State $\hat{z}_i^{(k-1)} \gets 1$ \label{lst:line:ln32}
            \vspace{0.5mm}
            \State $\mu \gets \mu - 1$ \label{lst:line:ln33}
        \EndIf
    \EndWhile \label{lst:line:ln35}
    \vspace{0.5mm}
    \State $\textbf{DeleteColumns}(\bm{X}^{(k)},\bm{X}_m^{(k-1)},\bm{\hat{z}}^{(k-1)}) = \bm{X}^{(k+1)},\bm{X}_m^{(k)},\bm{z}^{(k)}$ \label{lst:line:ln36}
    \vspace{0.5mm}
    \If {$\bm{z}^{(k-1)} = \bm{z}^{(k)}$} \label{lst:line:ln37}
    \vspace{0.5mm}
        \State $\beta \gets \beta - 1$ \label{lst:line:ln38}
    \EndIf
\EndWhile \label{lst:line:ln40}
\end{algorithmic}
\end{algorithm}

\section{Simulations}\label{sec:experiments}
In this study, we aim to compare the AFS-BM algorithm with the widely used feature selection approaches over the well-established real-life competition datasets. The efficacy of AFS-BM is demonstrated across both regression tasks, with Gradient Boosting Machines (GBMs) and Neural Networks (NNs) serving as the underlying models. Note that our algorithm is generic can be applied to other machine learning algorithms, however, we concentrate on these two since they are widely used in the literature.

\subsection{Implementation and Evaluation of AFS-BM}
Our experimental framework includes both synthetic and well-known competition datasets, and regression tasks. For regression analyses, first, we evaluate the performance of the AFS-BM algorithm compared to the other feature selection algorithms on the synthetic dataset. We start with a synthetic dataset to perform a controlled environment that allows us to thoroughly evaluate the algorithm's behavior and effectiveness in scenarios where the ground truth is known. This initial step provides valuable insights into the algorithm's capabilities before applying it to real-world datasets. Then, we continue with the real-life datasets including the M4 Forecasting Competition Dataset \citep{Makridakis} and The Istanbul Stock Exchange Hourly Dataset \citep{misc_istanbul_stock_exchange_247}. A crucial aspect of our study is performance benchmarking, where we compare the results of our AFS-BM feature selection algorithm against other prevalent methods in the field. Using a validation set, we optimize all hyperparameters for LightGBM, XGBoost, MLP, and the feature selection methods to remove any spurious or coincidental correlations that might arise from overfitting. The hyperparameter search space is also cross-validated.

\subsubsection{Regression Experiments and Results}
Our regression experiments use the daily series of the M4 Forecasting Competition Dataset \citep{Makridakis} and the Istanbul Stock Exchange Hourly Dataset \citep{misc_istanbul_stock_exchange_247}. We will first configure the AFS-BM algorithm for each dataset and then compare its performance against traditional feature selection methods. The experiments will also explore the impact of hyperparameter tuning on AFS-BM’s accuracy, concluding with insights into its overall effectiveness.

\paragraph{Time Series Feature Engineering and Algorithmic Evaluation on the Daily M4 Forecasting Dataset}
To predict the target value \( y_{t} \), we utilize the daily series from the M4 Forecasting Competition Dataset, which initially contains only the target values \( y_{t} \). For feature engineering, we consider the first three lags of \( y_{t} \) as \( y_{t-1} \), \( y_{t-2} \), and \( y_{t-3} \), and the 7th, 14th, and 28th lags as \( y_{t-7} \), \( y_{t-14} \), and \( y_{t-28} \) respectively. Beyond these basic lag features, we compute rolling statistics, specifically, the rolling mean and rolling standard deviation with window sizes of 4, 7, and 28 until \(y_t\).

To remove any bias to a particular time series, we select 100 random series from the dataset, normalize them to $[-1,1]$ after mean removal. Since the data length is not uniform for each series, we reserve the last $10\% $ of total samples for testing, $20\% $ of total samples before the testing as mask validation, and $20\% $ of total samples before the mask validation as model validation. To underscore the significance of feature selection, we utilize the default configurations of LightGBM, XGBoost, and Multi-Layer Perceptron (MLP) without incorporating any specific feature selection techniques. We compare our algorithm against the well-known feature selection methods, Cross-Correlation, Mutual Information, and RFE. Using a validation set, we optimize all hyperparameters for LightGBM, XGBoost, MLP, and the feature selection methods. Both our introduced AFS-BM algorithm and the other feature selection methods employ LightGBM and XGBoost as their primary boosting techniques.

\begin{figure}[ht]
    \centering
    \includegraphics[width=0.50\textwidth,left]{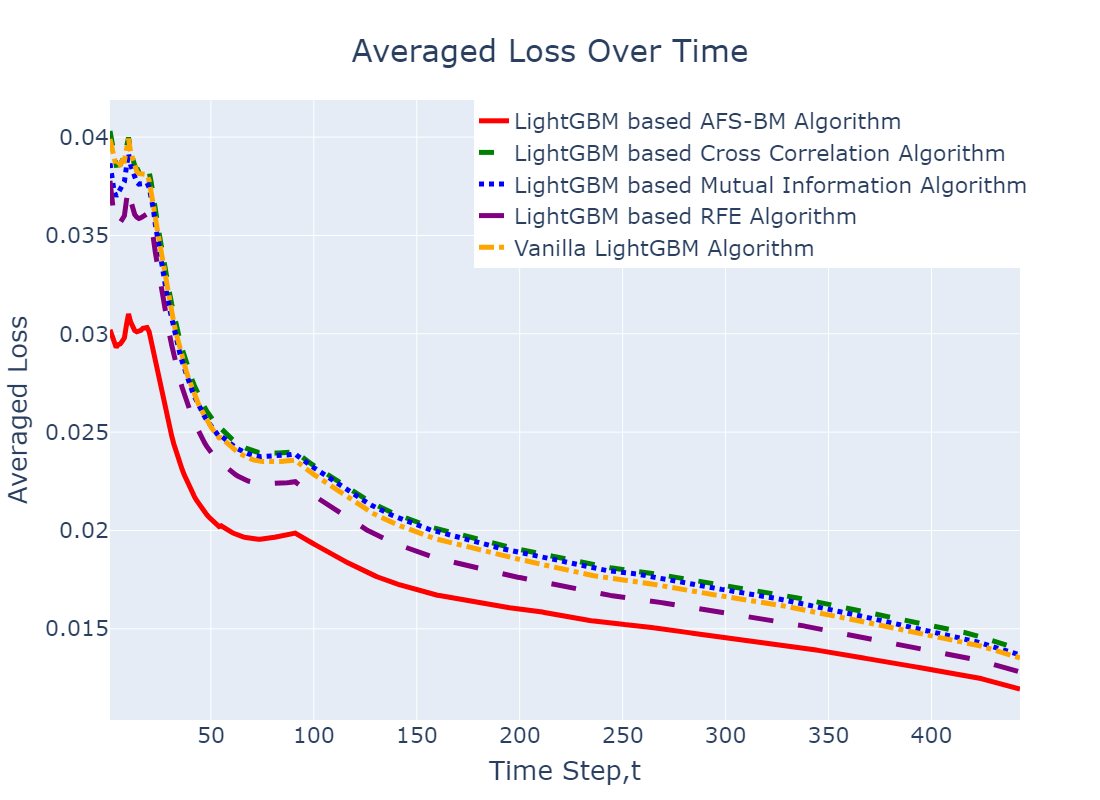}
    \caption{Comparison of averaged loss over time, $l_t^{(ave2)}$, for the experiments on M4 Competition data with LightGBM-based algorithms.}
    \label{fig:m4_lgbm}
\end{figure}

\begin{figure}[ht]
    \centering 
    \includegraphics[width=0.50\textwidth,left]{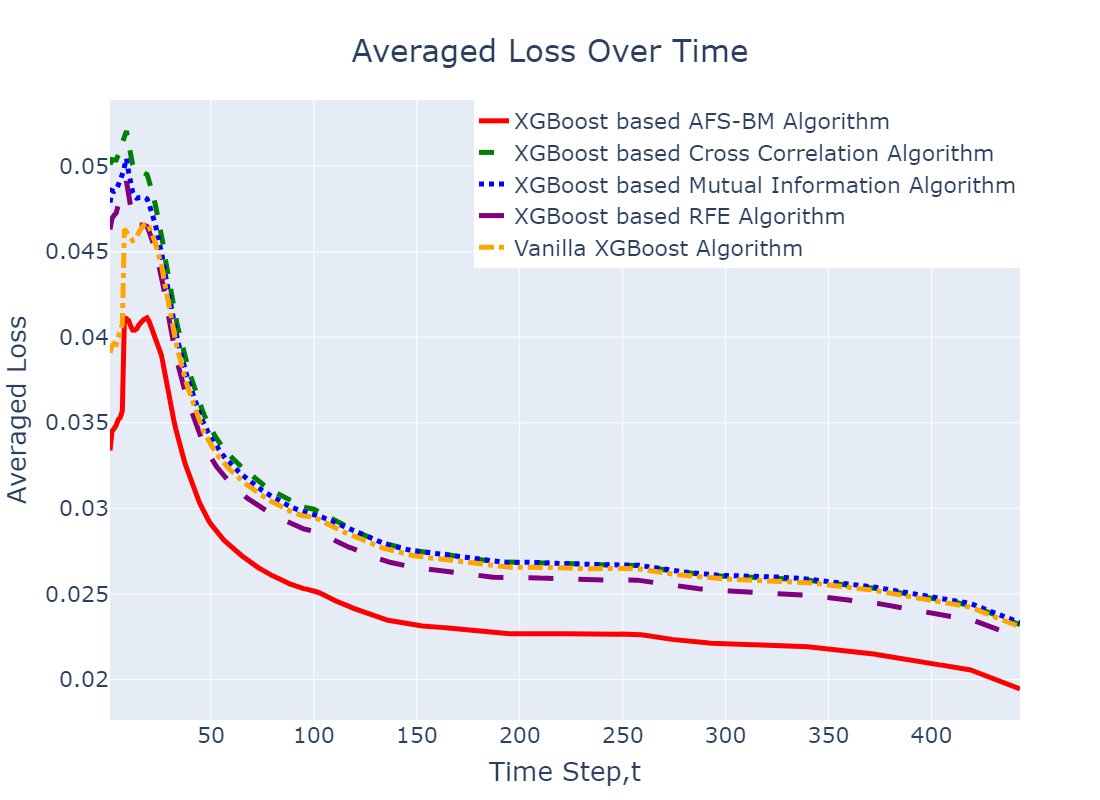}
    \caption{Comparison of averaged loss over time for the experiments on M4 Competition data with XGBoost-based algorithms.}
    \label{fig:m4_xgb}
\end{figure}

For each randomly selected time series, denoted as \(y_t^{(s)}\), where \(s = 1, \ldots, 100\), we apply all the algorithms for feature selection and then compute the test loss. The test loss are given by \(l_t^{(s)}\). Here, \(l_t^{(s)} \triangleq (y_t^{(s)} - \hat{y}^{(s)}_{t})^2\), and \(t\) ranges from 1 up to \(t_{max}\), the latter being the longest time duration among all \(y_t^{(s)}\) for \(k = 1, \ldots, 100\). For datasets with shorter durations, we pad the loss sequences with zeros at the end to ensure consistency. To eliminate any bias towards a specific sequence, we compute the average of these loss sequences, yielding the final averaged squared error: \(l_t^{(ave)} = \frac{1}{100} \sum_{s=1}^{100} l_t^{(s)}\). For further refinement, we also average over time, resulting in \(l_t^{(ave2)} = \frac{1}{t} \sum_{j=1}^t l_j^{(ave)}\). The outcomes for the LightGBM-based algorithms are depicted in Fig. \ref{fig:m4_lgbm}, while those for the XGBoost-based algorithms are illustrated in Fig. \ref{fig:m4_xgb}. 

Additionally, Table \ref{tab:1} offers a detailed comparison of our algorithm against the standard LightGBM, XGBoost, and their respective implementations with Cross-Correlation, Mutual Information, and RFE algorithms.

\begin{table}[ht]
\centering

\resizebox{0.5\textwidth}{!}{\begin{tabular}[]{cccc}
\hline
Algorithm\textbackslash Base Model        & LightGBM                  & XGBoost   \\ \hline
Cross-Correlation Algorithm           & \SI{4.7895e-2}{}          & \SI{9.1995e-2}{}            \\
Mutual Information Algorithm          & \SI{4.6626e-2}{}          & \SI{9.0114e-2}{}            \\
RFE Algorithm                         & \SI{4.5212e-2}{}          & \SI{8.8511e-2}{}            \\
Vanilla Algorithm                     & \SI{4.7691e-2}{}          & \SI{8.9000e-2}{}        \\
AFS-BM Algorithm                      & $\bm{3.8251\times10^{-2}}$& $\bm{5.4094\times10^{-2}}$ \\ \hline
\end{tabular}}
\label{tab:1}
\caption{Final MSE Results of the Experiments on M4 Competition Daily Dataset with GBM-based Learners}
\end{table}

\begin{figure}[ht]
    \centering
    \includegraphics[width=0.50\textwidth,left]{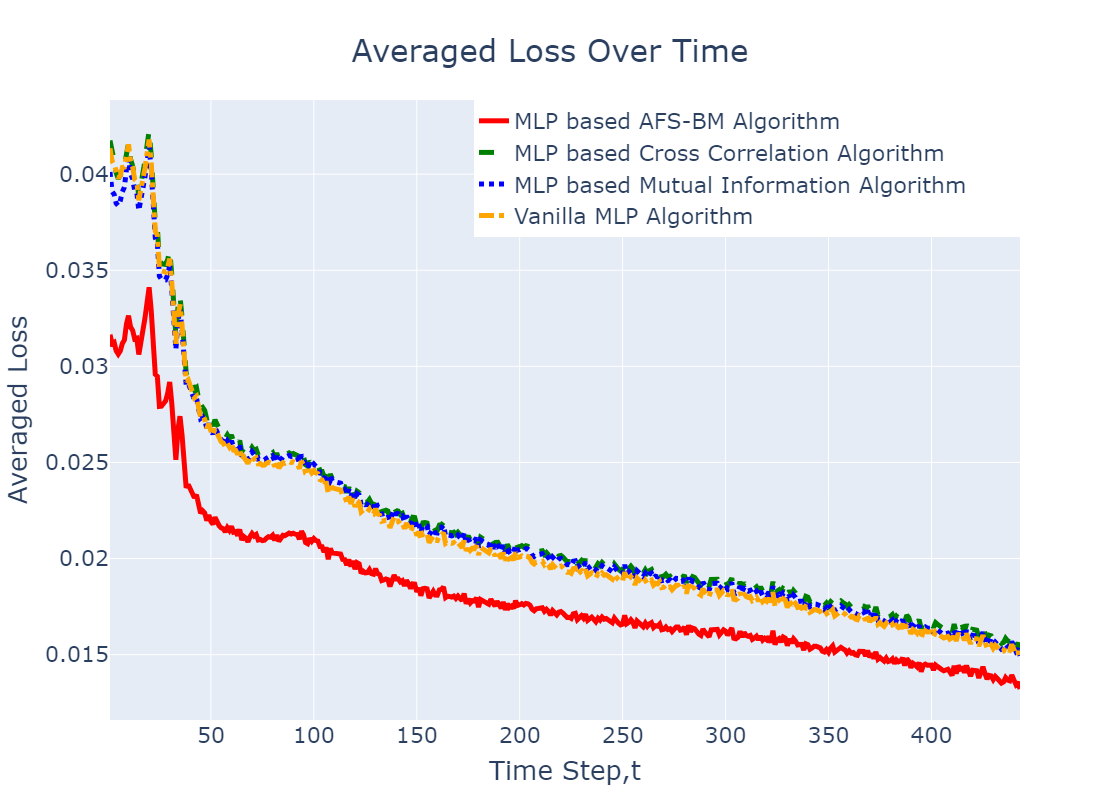}
    \caption{Comparison of averaged loss over time for the experiments on M4 Competition data with MLP-based algorithms.}
    \label{fig:m4_mlp}
\end{figure}

Moreover, we apply the MLP model and other algorithms to the test data for each series, resulting in the loss sequences. The performance trends of the algorithms over time are visualized in Figure \ref{fig:m4_mlp}, which provides a graphical representation of the averaged loss sequences for the experiments on the M4 Competition data with MLP-based algorithms.

Additionally, Table \ref{tab:mlp_m4} offers a detailed comparison of our algorithm against the standard MLP and their respective implementations with Cross-Correlation, Mutual Information, and RFE algorithms.

\begin{table}[ht]
\centering
\resizebox{0.45\textwidth}{!}{\begin{tabular}[]{cccc}
\hline
Algorithm\textbackslash Base Model        & MLP   \\ \hline
Cross-Correlation Algorithm           & \SI{4.8052e-2}{}            \\
Mutual Information Algorithm          & \SI{4.6781e-2}{}            \\
Vanilla Algorithm                     & \SI{4.7845e-2}{}        \\
AFS-BM Algorithm                      & $\bm{3.8385\times10^{-2}}$ \\ \hline
\end{tabular}}
\label{tab:mlp_m4}
\caption{Final MSE Results of the Experiments on M4 Competition Daily Dataset with MLP}
\end{table}
\vspace{-10mm}
\paragraph{Time Series Feature Engineering and Algorithmic Evaluation on the Istanbul Stock Exchange Dataset}

To forecast the target value \( y_{t} \), we turn to the Istanbul Stock Exchange Dataset\citep{misc_istanbul_stock_exchange_247}, which provides hourly data points. Initially, this dataset offers only the target values, denoted as \( y_{t} \). To enrich the predictive capability of our model, we perform a systematic feature engineering process. Firstly, we extract immediate temporal dependencies by considering the three most recent lags of \( y_{t} \), represented as \( y_{t-1} \), \( y_{t-2} \), and \( y_{t-3} \). Recognizing the significance of longer-term patterns in hourly data, especially over the course of a day or multiple days, we also incorporate the 24th, 36th, and 48th lags, denoted as \( y_{t-24} \), \( y_{t-36} \), and \( y_{t-48} \) respectively. In addition to these lags, we also factor in the cyclical nature of time by taking the sine and cosine values of the timestamp, which includes the month, day, and hour. This helps in capturing the cyclic patterns associated with daily and monthly rhythms.

In addition to these features, we also calculate the rolling statistics to capture more nuanced patterns. Specifically, we calculate the rolling mean and rolling standard deviation for both recent (\( y_{t-1} \)) and daily (\( y_{t-24} \)) lags. These calculations are performed over window sizes of 4 (representing a shorter part of the day), 12 (half a day), and 24 (a full day) hours. Hence, we get a comprehensive set of features from the time series data.

Following mean removal, the dataset is first normalized to $[-1,1]$. Given the dataset, $10\%$ of total samples are designated for testing. Furthermore, $20\%$ of the samples leading up to the test set are reserved for mask validation, while an additional $20\%$ preceding the mask validation is allocated for model validation. LightGBM, XGBoost, and MLP are utilized in their standard configurations, bypassing specialized feature selection methods. Our algorithm is compared with the algorithms that are the same as the previous test sets. All hyperparameters associated with LightGBM, XGBoost, and MLP undergo optimization based on the model validation set to ensure optimal performance.

The findings based on Mean Squared Error (MSE) loss and the number of selected features for both LightGBM and XGBoost-based algorithms are displayed in Table \ref{tab:2} and Table \ref{tab:ise_features}. 
\begin{table}[ht]
\centering
\resizebox{0.5\textwidth}{!}{\begin{tabular}[]{cccc}
\hline
Algorithm\textbackslash Base Model        & LightGBM                  & XGBoost   \\ \hline
Cross-Correlation Algorithm           & \SI{2.5564e-2}{}          & \SI{2.5876e-2}{}            \\
Mutual Information Algorithm          & \SI{2.8480e-2}{}          & \SI{2.5536e-2}{}            \\
RFE Algorithm                         & \SI{4.5903e-3}{}          & \SI{5.8105e-3}{}            \\
Vanilla Algorithm                     & \SI{4.3225e-3}{}          & \SI{5.5866e-3}{}        \\
AFS-BM Algorithm                      & $\bm{3.8350\times10^{-3}}$& $\bm{3.9445\times10^{-3}}$ \\ \hline
\end{tabular}}
\caption{Final MSE Results of the Experiments on Istanbul Stock Exchange Hourly Dataset with GBM-based Learners}
\label{tab:2}
\end{table}

\begin{table}[ht]
    \centering
    \resizebox{0.45\textwidth}{!}{
    \label{tab:ise_features}

    \begin{tabular}[]{cccc}

         Base Model & Method & Selected Features \\
\midrule
         LightGBM & AFS-BM &2 \\
         XGBoost & AFS-BM &8  \\
         LightGBM & Cross-Correlation &66 \\
         XGBoost & Cross-Correlation &66   \\
         LightGBM & Mutual Information &97   \\
         XGBoost & Mutual Information &101   \\
         LightGBM & RFE &10   \\
         XGBoost & RFE &5   \\
\bottomrule
    \end{tabular}}
    \caption{The number of selected features selected from a total of 176 features by each feature selection method on the Istanbul Stock Exchange Hourly Dataset with GBM-based Learners}
\end{table}
\vspace{-5mm}
Furthermore, the dataset undergoes evaluation using the MLP model. MLP, with its layered architecture, is adept at capturing nonlinear relationships in the data, making it a suitable choice for such complex datasets. The results based on Mean Squared Error (MSE) loss for the MLP-based algorithms are presented in Table \ref{tab:mlp_istanbul}. 

\begin{table}[ht]
\centering
\resizebox{0.45\textwidth}{!}{
\begin{tabular}[]{cccc}
\hline
Algorithm\textbackslash Base Model        & MLP   \\ \hline
Cross-Correlation Algorithm           & \SI{2.5652e-2}{}            \\
Mutual Information Algorithm          & \SI{2.8603e-2}{}            \\
Vanilla Algorithm                     & \SI{4.3350e-3}{}        \\
AFS-BM Algorithm                      & $\bm{3.8423\times10^{-3}}$ \\ \hline
\end{tabular}}
\label{tab:mlp_istanbul}
\caption{Final MSE Results of the Experiments on Istanbul Stock Exchange Hourly Dataset with MLP}
\end{table}

A comparative analysis reveals that the AFS-BM algorithm consistently surpasses other methods across all regression experiments. The main strength of AFS-BM is its ability to adapt its feature selection as data patterns change. It does this by checking the error at every step. This adaptability is different from fixed methods such as Cross-Correlation and Mutual Information. AFS-BM can effectively handle both short-term and long-term changes in time series data. When combined with cross-validated hyperparameter selection, the algorithm works at its best. In summary, our experimental results demonstrate the superior performance of AFS-BM algorithm over the state-of-the art feature selection methods.

\section{Conclusion} \label{sec:conclusion}
In this paper, we have addressed the critical problem of feature selection in general machine learning models, recognizing its critical role in both model accuracy and efficiency. Traditional feature selection methods struggle with significant challenges such as scalability, managing high-dimensional data, handling correlated features, adapting to shifting feature importance, and integrating domain knowledge. To this end, we introduced the ``Adaptive Feature Selection with Binary Masking" (AFS-BM) algorithm.

AFS-BM differs from the current algorithms since it uses joint optimization for both dynamic and stochastic feature selection and model training, a method that involves binary masking. This approach enables the algorithm to adapt continuously by adjusting features and model parameters during training, responding to changes in feature importance measured by the loss metric values. This adaptability ensures that AFS-BM retains essential features while discarding less relevant ones based on evaluation results, rather than relying solely on feature importance calculated using intuitive reasoning. This is crucial since depending solely on feature importance can lead to the removal of informative features, as it evaluates them within the context of the feature set, rather than considering their contribution to final model performance. AFS-BM's approach prevents the removal of informative features, ultimately enhancing model accuracy. To encourage further research and allow others to replicate our results, we openly share our source code\footnote{\url{https://github.com/YigitTurali/AFS_BM-Algorithm}}. 
\section*{Statements and Declarations}
\begin{itemize}
\item Competing interests: The authors declare that they have no known competing financial interests or personal relationships that could have influenced influence the work reported in this paper.
\item Availability of data and materials : The data that support the findings of this study is openly available in UCI Machine Learning Repository at \url{https://archive.ics.uci.edu}.

\end{itemize}

% \begin{appendices}

% \section{Section title of first appendix}\label{secA1}

% An appendix contains supplementary information that is not an essential part of the text itself but which may be helpful in providing a more comprehensive understanding of the research problem or it is information that is too cumbersome to be included in the body of the paper.




% \end{appendices}

\bibliographystyle{achemso}
\bibliography{sn-bibliography}% common bib file

\end{document}